\documentclass{article}
\usepackage{spconf,amsmath,graphicx}

\usepackage{algorithm}
\usepackage{algorithmic}
\usepackage{multirow}
\usepackage{amsfonts}
\usepackage{booktabs}

\usepackage{hyperref}

\title{Expression Domain Translation Network \\ for Cross-domain Head Reenactment}
\name{Taewoong Kang$^{1,\ast}$\, Jeongsik Oh$^{2,\ast}$\, Jaeseong Lee$^{2}$\, Sunghyun Park$^{2}$\, Jaegul Choo$^{2}$}
\address{$^1$ Korea University \qquad $^2$ KAIST}
\begin{document}
%
\maketitle

\begin{abstract}

Despite the remarkable advancements in head reenactment, the existing methods face challenges in cross-domain head reenactment, which aims to transfer human motions to domains outside the human, including cartoon characters. 
It is still difficult to extract motion from out-of-domain images due to the distinct appearances, such as large eyes.
Recently, previous work introduced a large-scale anime dataset called AnimeCeleb and a cross-domain head reenactment model, including an optimization-based mapping function to translate the human domain's expressions to the anime domain. 
However, we found that the mapping function, which relies on a subset of expressions, imposes limitations on the mapping of various expressions. 
To solve this challenge, we introduce a novel \textit{expression domain translation network} that transforms human expressions into anime expressions.
Specifically, to maintain the geometric consistency of expressions between the input and output of the expression domain translation network, we employ a \textit{3D geometric-aware loss function} that reduces the distances between the vertices in the 3D mesh of the human and anime.
By doing so, it forces high-fidelity and one-to-one mapping with respect to two cross-expression domains. 
Our method outperforms existing methods in both qualitative and quantitative analysis, marking a significant advancement in the field of cross-domain head reenactment.
  
\end{abstract}


\section{Introduction}

\begin{figure}[t!]
    \centering
    \includegraphics[width=1.0\linewidth]{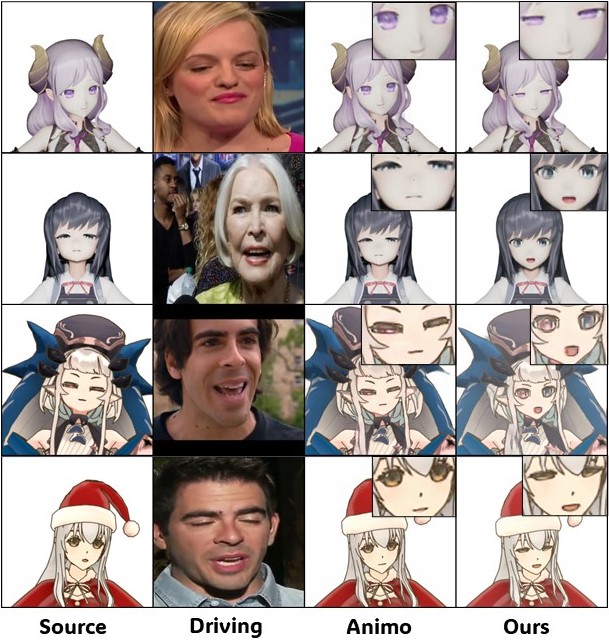}
    \caption{Cross-domain head reenactment examples of our method. Given the anime image, we edit it by injecting the pose and expression from driving image.}
    \label{fig:tearue}
    \vspace{-0.6cm}
\end{figure}

Given the advancements in online live streaming platforms such as YouTube, there has been a growing trend among users to express themselves through virtual avatars (\textit{i.e.}, virtuber).
This trend has elevated the significance of head reenactment tasks, wherein human motions are transferred to other virtual characters in response to such user needs.
Recent studies~\cite{firstorder, onefreeview, ren2021pirenderer, drobyshev2022megaportraits, zhang2023metaportrait} in head reenactment have made it possible to transfer human motions onto other human heads, by leveraging large-scale datasets of human talking head videos.
However, existing head reenactment methods still face challenges when applied to virtual avatars, such as anime characters, which exist outside the human domain.

Such a process of transferring human head motion to images from domains outside the human is referred to as cross-domain head reenactment.
Conventional head reenactment approaches typically encode the human motion through landmarks~\cite{firstorder, onefreeview}, 3D morphable models (3DMM)~\cite{ren2021pirenderer}, or latent descriptors~\cite{burkov2020neural}.
However, extracting motion from out-of-domain images (\textit{e.g.}, cartoon characters) with distinct appearances, including small mouths or large eyes, is challenging.
Moreover, due to the shortage of suitable video datasets for head reenactment for other domains, it is difficult to facilitate cross-domain head reenactment.

To address these challenges, only a few works~\cite{kim2022animeceleb, bansal2018recycle, song2021everything, xu2022motion} have attempted to tackle cross-domain head reenactment.
Recently, AnimeCeleb~\cite{kim2022animeceleb} has endeavored to facilitate head reenactment of cartoon characters by constructing a large-scale animation dataset based on 3D character models, including pairs of 2D cartoon images and pose vectors.
Furthermore, they proposed a cross-domain head reenactment approach leveraging two domain datasets (\textit{i.e.}, AnimeCeleb and VoxCeleb~\cite{nagrani2017voxceleb}).
Specifically, AnimeCeleb has designed an optimization-based mapping function to transform AnimeCeleb's pose vectors into the 3DMM space, leveraging landmarks of specific expressions (\textit{e.g.}, left closed eye).
Even when trained solely on AnimeCeleb, the model is applicable to a wide range of cartoon images with various styles, such as Waifu Labs~\footnote{https://waifulabs.com/}, Naver Webtoon~\footnote{https://comic.naver.com/} and 2D Disney~\footnote{https://toonify.photos/\vspace{-1.2cm}}.
However, we discovered that during the process of mapping expressions of AnimeCeleb's pose vector to the 3DMM spaces, the optimization-based method relying on only a subset of expressions imposes limitations on mapping various expressions, as illustrated in Fig.~\ref{fig:tearue}.

To overcome the lack of reflecting facial expressions caused by the limitations, we propose a novel \textit{expression domain translation network} to map the human expressions to the anime expressions.
This network is designed to transform 3DMM parameters into semantically equivalent pose vectors.
Specifically, our main idea is to train the network with our novel \textit{3D geometric-aware loss} that reduces the distances between the vertices in the 3D mesh of the human and anime.

This approach aims to maintain the geometric consistency of the two different domain's expressions utilizing the shared vertex space.
By doing so, it forces high-fidelity and one-to-one mapping with respect to two cross-expression domains. 
Through experiments, we demonstrate the superiority of our method in the field of cross-domain talking heads.

\section{Methodology}

In this section, we present our \textit{cross-domain head reenactment framework}. 
Given the driving image $\mathbf{I}_d$ of the human domain with the 3DMM vector $\mathbf{p}$, our model aims to generate anime image $\hat{\mathbf{I}}$ by modifying the head pose and face expressions of the source image $\mathbf{I}_s$ of the anime domain.

\subsection{Expression Space Domain Gap }

We employ a DECA~\cite{feng2021learning} as a 3DMM encoder to extract FLAME~\cite{li2017learning} parameters to encode human motion.
Specifically, with FLAME parameters, human face mesh can be represented as:
\begin{equation}\label{eq1}
    \mathbf{T}_P(\beta,\theta,\psi)=
    \mathbf{T}+B_S(\beta;S)+B_P(\theta;P)+B_E(\psi;E),
\end{equation}
where $\mathbf{B}_S(\beta;S)$, $B_P(\theta;P)$, and $B_E(\psi;E)$ denote shape, head angle, and expression blendshapes, and $\mathbf{T}$ indicates the average mesh shape in zero pose.
However, we only utilize expression coefficients $\psi\in\mathbb{R}^{50}$ and head pose $\theta \in \mathbb{R}^{6}$, where $\theta = [pose; jaw];pose\in\mathbb{R}^{3};jaw\in\mathbb{R}^{3}$.

Basically, we leverage the AnimeCeleb dataset~\cite{kim2022animeceleb}, which consists of pairs of anime images and pose vectors.
While for encoding anime motion, we utilize AnimeCeleb's pose vectors $\mathbf{v}\in\mathbb{R}^{20}$, which consist of 17-dimensional expression coefficients $\mathbf{b}\in\mathcal{B}$ and head angles $\mathbf{h}\in\mathcal{H}$, following the previous work~\cite{kim2022animeceleb}. 
Specifically, expression coefficients $\mathbf{b}$ consist of six eye-related dimensions, six eyebrow-related dimensions, and five mouth-related dimensions.
For instance, the first dimension of the pose vector $\mathbf{v}$ corresponds to a left-eye wink, and thus it holds a value within the range of 0 to 1, varying according to the degree of the eye being closed.

\begin{figure*}[t!]
    \centering
    \includegraphics[width=1.0\linewidth]{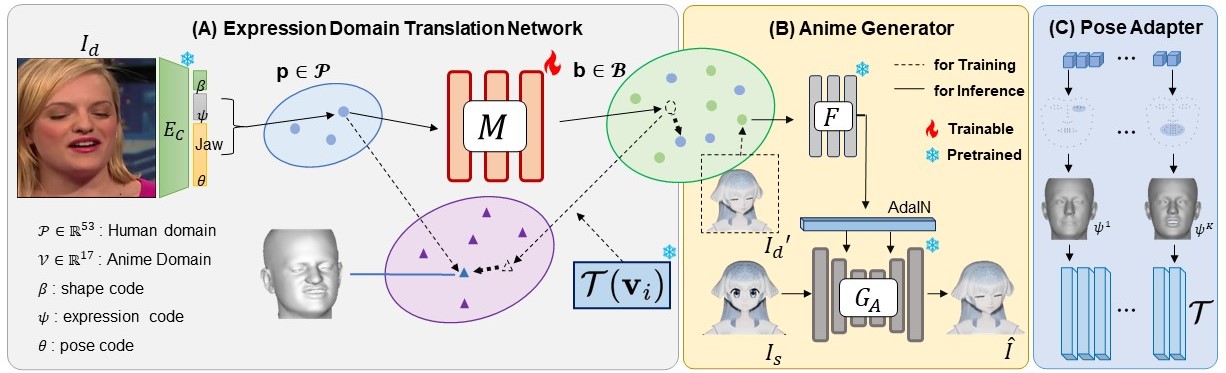}
    \vspace{-0.7cm}
    \caption{Overview of our framework. For visual clarity, we have omitted the head rotation parameters.}
    \label{fig:model}
    \vspace{-0.4cm}
\end{figure*}

There exist differences in the representation of facial pose and expression between human and anime.
Therefore, for effectively transferring human motion to anime images, it is crucial to accurately map FLAME parameters encoding human motion to pose vectors representing anime motion.
Notably, while head angles allow for precise one-to-one mapping as described in the previous work~\cite{kim2022animeceleb}, the approach for translating human's expression coefficients $\psi\in\mathbb{R}^{50}$ into anime's expression coefficients $\mathbf{b}\in\mathbb{R}^{17}$ is required.

\subsection{Expression Domain Translation Network}

To map the human's expression coefficients $\psi$ into anime's expression coefficients $\mathbf{b}$, we propose \textit{expression domain translation network} $\mathbf{M}$.
Specifically, with $p =[\psi; jaw]$ as input variables that are related to the face expressions, the model outputs the expression coefficients $\mathbf{b}$.
Given that FLAME incorporates expressions in conjunction with the jaw to construct the mesh, the use of $jaw$ becomes an indispensable component.
When we train the network, a paired dataset that denotes equivalent facial expressions on two different domains is required.
However, due to the domain discrepancy, it becomes imperative to train the model using an unpaired dataset.
To address this issue, we utilize a \textit{pose adapter} and train the expression domain translation network $\mathbf{M}$ with a \textit{3D geometric-aware loss}.

\noindent\textbf{Pose Adapter.} 
To map the anime's expression coefficients $\mathbf{b}$ onto the vertex space, it needs to be converted into $\psi$ form.
Consequently, we employed the use of a pose adapter $\mathcal{T}$.
In order to obtain $\mathcal{T}$, we use the same step with mapping function $\mathcal{T}$ from Animo~\cite{kim2022animeceleb}. 
The difference between the mapping function and the pose adapter is that the mapping function maps to BFM~\cite{gerig2018morphable} expression parameters $\beta\in\mathbb{R}^{64}$, while the pose adapter maps to FLAME~\cite{li2017learning} expression parameters $\psi\in\mathbb{R}^{50}$.
Moreover, unlike the mapping function, which uses the mapping anime to human expression directly, we only use a pose adapter to send it to the vertex space for the purpose of mapping the human to anime expression more continuously.

\noindent\textbf{3D Geometric-Aware Loss.}
To identify the anime’s expression coefficients $\mathbf{b}$ that is semantically equivalent to $\psi$, we train the expression translation network $\mathbf{M}$ utilizing 3D geometric-aware loss.
Due to the expression space domain gap, we map both $\mathbf{b}$ and $\psi$ to the vertex space, which is a mutually compatible space.
Additionally, utilizing the vertex space allows us to use geometrically aware information.
Exploiting these characteristics, we train the parameters from each domain to possess expressions that are geometrically congruent. 
To ensure that the corresponding vertices have identical coordinates, we employ a vertex loss term.
This term applies mean squared error loss between the predicted and actual 3D coordinates of each vertex constituting the mesh.
Additionally, we extract just the 68 keypoints $\mathbf{k}_i$ from vertices to further train on important information.
\begin{equation}\label{eq4}
    \mathcal{L}_{lm} = \sum_{i=1}^{68}\parallel \hat{\mathbf{k}}_i-\mathbf{k}_i \parallel_1.
\vspace{-0.2cm}
\end{equation}
To better capture sensitive and important features like the eyes and mouth, we have applied an eye and mouth closure loss.
This loss computes the relative offset of landmarks
$\hat{\mathbf{k}}_i$ and $\hat{\mathbf{k}}_j$ 
on the upper and lower eyelid and outer mouth, and measures the difference to the offset of the corresponding predicted landmarks
$\mathbf{k}_i$ and $\mathbf{k}_j$.
The loss is defined as
\begin{equation}\label{eq5}
    \mathcal{L}_{eye, mouth} = \sum_{(i,j) \in E,M} \parallel |\hat{\mathbf{k}}_i-\hat{\mathbf{k}}_j| - |\mathbf{k}_i-\mathbf{k}_j| \parallel_1,
\vspace{-0.2cm}
\end{equation}
where $E$ is the set of upper and lower eyelid landmark pairs and $M$ is the set of upper/lower outer mouth landmark pairs. 
In summary, our full objective function is given as:
\begin{equation}\label{eq2}
    \mathcal{L}_{total} = \mathcal{L}_{lm} + \mathcal{L}_{eye, mouth} + \lambda_{ver}\cdot\mathcal{L}_{ver}.
\vspace{-0.2cm}
\end{equation}
Here, $\lambda_{ver}$ is the hyperparameter and set to 100.

\subsection{Anime Generator}

Fig.~\ref{fig:model} provides an overview of our framework.
In this section, we introduce the remaining part of our framework, which synthesizes anime images based on the pose vectors predicted from the expression domain translation network.
Our generator is based on PIRenderer~\cite{ren2021pirenderer}, following previous work~\cite{kim2022animeceleb}.

\noindent\textbf{Motion Network.} 
With a driving pose $\mathbf{v}$, the motion network $F$ generates a latent pose code $z$.
Thanks to the $\mathbf{M}$ and the characteristics of $\mathbf{v}$, the motion network $F$ can be designed as the domain-agnostic and controllable method, which is the main difference with PIRender~\cite{ren2021pirenderer}.
Then, we just need the generator that can edit the source image with the given $z$.

\noindent\textbf{Warping \& Editing Network.} 
With warping and editing network, we can generate an image that is guided by $z$ through adaptive instance normalization (AdaIN)~\cite{huang2017arbitrary}.
A warping network predicts the optical flow $\mathbf{u}$ that serves to approximate the coordinate offsets to reposition a source head like a driving head.
An editing network that serves to portray a detailed expression-related pose gets the source image, optical flow $\mathbf{u}$, and latent pose code $z$.
Refer to PIRenderer~\cite{ren2021pirenderer} for details.

With our expression domain translation network and the anime generator, we are capable of achieving state-of-the-art performance in cross-domain head reenactment.

\section{Experiments}

\subsection{Experiment Setup}

\noindent\textbf{Datasets.}
To train our expression domain translation network \textbf{M}, we take a subset of videos from Voxceleb~\cite{nagrani2017voxceleb}. We downloaded 18,503 videos for train set and 504 videos for test set. Also, we use AnimeCeleb~\cite{kim2022animeceleb} dataset to train anime generator. 

\noindent\textbf{Training Details.}
The expression domain translation network \textbf{M} and anime generator \textbf{G} are trained separately.
For the expression domain translation network \textbf{M}, we trained it for 50 epochs, where the batch size is 512, and the optimizer is Adam with a learning rate of $1\times10^{-4}$.
For the generator, we trained the model for 200 epochs, where the batch size is 8, and the optimizer is Adam with a learning rate of $1\times10^{-4}$.

\subsection{Comparison with Baselines}
We compare out model with the baselines such as FOMM~\cite{firstorder}, PIRenderer+$\mathcal{T}$~\cite{ren2021pirenderer,kim2022animeceleb}, and Animo~\cite{kim2022animeceleb}. 
Moreover, we have empirically substantiated the efficacy of our model with respect to its distributional characteristics.

\noindent\textbf{Quantitative Evaluation.}
Table~\ref{Table:quantity} shows quantitative comparisons between our model and the baselines~\cite{firstorder, ren2021pirenderer, kim2022animeceleb} on the cross-domain face reenactment.
When evaluating cross-domain face reenactment, we found that existing metrics do not adequately capture facial expressions. 
Therefore, we introduce a new metric, called the Keypoint Distance Ratio (KDR), which measures the $\ell_1$  distance ratio of eyes compared with neutral keypoints' distance. 
The reason why we compare the relative distance of eyes is that there is topological heterogeneity between human and anime character domain (\textit{e.g.,} Anime character's abstract distance of eye's lid is innately larger than humans'). 
Specifically, this ratio compares the upper and lower eyelid distances in both the driving image and the predicted image.
For a more detailed analysis, we employ a keypoint detector~\cite{anime-face-detector} designed for the anime domain.
Because the keypoints in anime are different from those in humans, we measure the $\ell_1$ distance and compare ratios, considering the geometrical heterogeneity.
As evidenced in Table~\ref{Table:quantity}, our model outperforms in both FID and KDR metrics despite its advantages of a smaller size and shorter training time due to single-dataset training.

\begin{figure}[t!]
    \centering
    \includegraphics[width=1.0\linewidth]{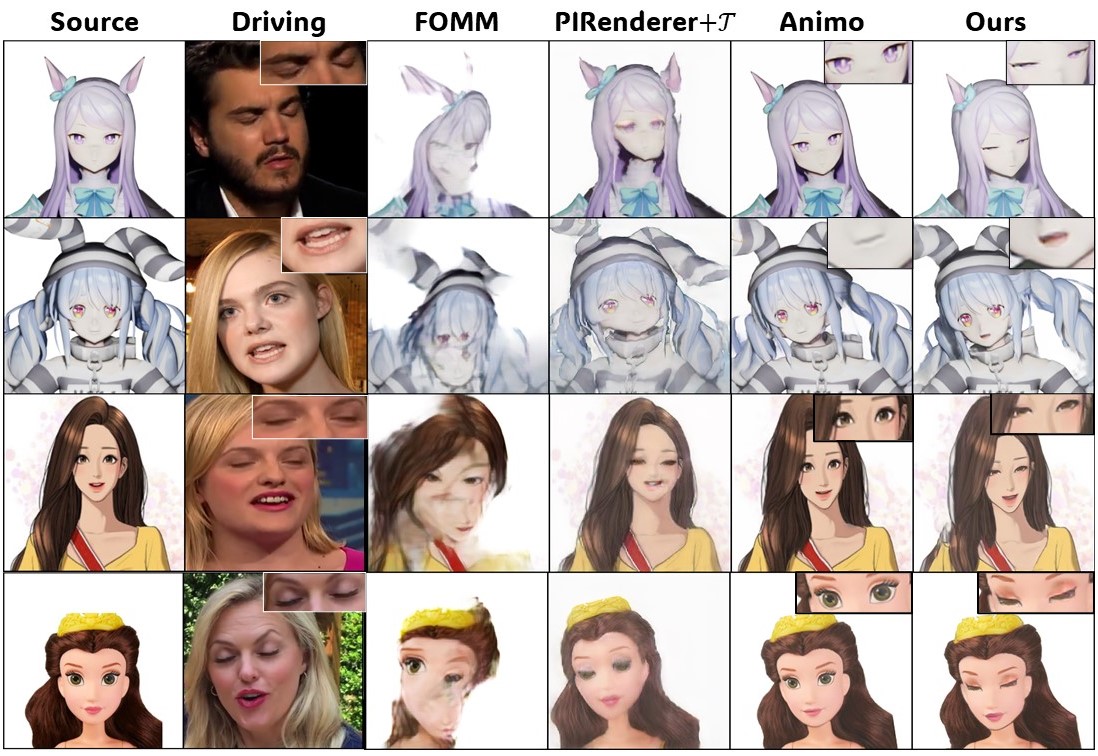}
    \caption{Qualitative comparison between our model and the baselines on cross-domain face reenactment given the source image from AnimeCeleb~\cite{kim2022animeceleb}, Naver Webtoon, 2D Disney and the driving image from VoxCeleb~\cite{nagrani2017voxceleb}}
    \label{fig:quality}
    \vspace{-0.3cm}
\end{figure}

\begin{figure}[t!]
    \centering
    \includegraphics[width=1.0\linewidth]{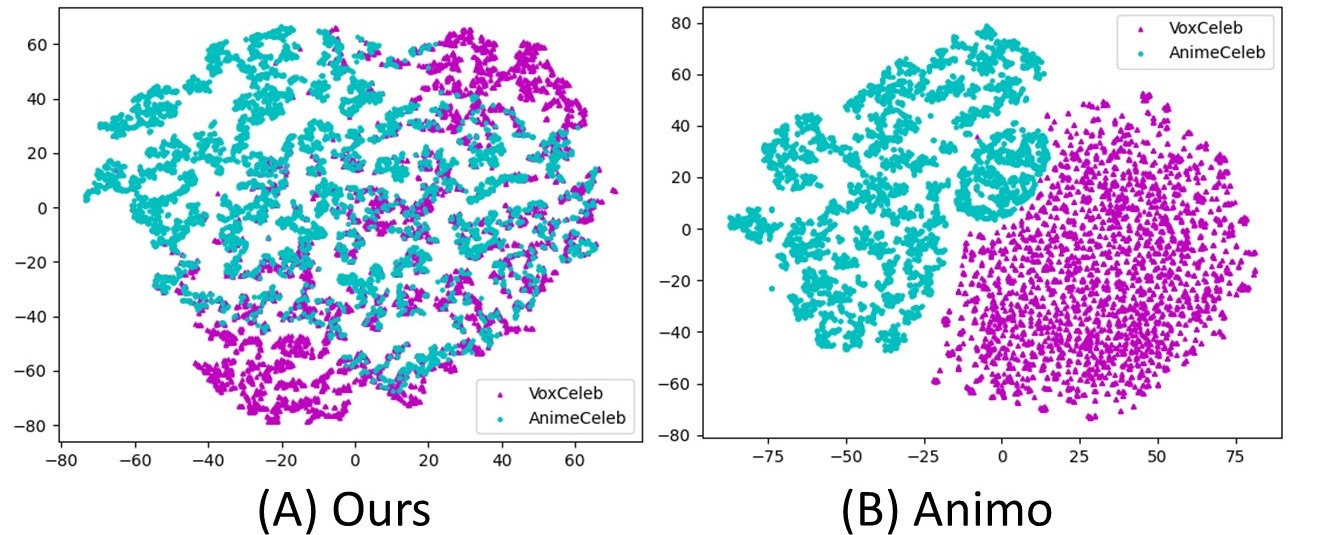}
    \caption{T-SNE visualization. (A) Ours shows a lower expression distribution discrepancy than (B) Animo.}
    \label{fig:t-sne}
    \vspace{-0.7cm}
\end{figure}

\noindent\textbf{Qualitative Evaluation.}
Fig.~\ref{fig:quality} shows qualitative comparisons between our model and the baselines.
In the realms of texture and identity preservation, it is discernible that both our method and Animo~\cite{kim2022animeceleb} clearly outperform FOMM~\cite{firstorder} and PIRenderer+$\mathcal{T}$~\cite{ren2021pirenderer}.
Moreover, while Animo exhibits limited capability in faithfully capturing facial expressions, our method demonstrates a markedly superior performance in accurately reflecting them.
We conclude that the expression domain translation network \textbf{M} serves as an instrumental component, facilitating the successful cross-domain transfer of facial expressions within the model.
More qualitative results and video results are available on the project website~\footnote{\href{https://keh0t0.github.io/research/EDTN/}{https://keh0t0.github.io/research/EDTN/} \vspace{-0.6cm}}.

\noindent\textbf{Distribution Concurrency.}
One of the key factors contributing to the efficacy of our method is the successful alignment of distributions across both domains. 
To empirically validate this, we conducted a t-SNE analysis on a sample size of 5,000 data points within the input space of the motion network.
The results can be visualized in Fig.\ref{fig:t-sne}.

\subsection{Ablation Study}
To empirically substantiate the necessity of our loss function, we conducted a comprehensive ablation study by dropping the loss component.
As evident from Table \ref{Table:talkinghead}, the model that was subjected to the full complement of loss terms demonstrated the highest performance in KDR.
Moreover, even the configuration that was trained solely with the vertex loss outperformed Animo~\cite{kim2022animeceleb}, thereby substantiating the efficacy of imposing loss constraints within the vertex space.

\begin{table}[t!]
    \small
    \centering
    \begin{tabular}{l@{}cccccc}
    \toprule
    \multirow{2}{*}{\textbf{Train Dataset}} &
    \multirow{2}{*}{\textbf{Model}} & 
    \multicolumn{2}{c}{\textbf{Cross-Domain}} \\
    \cmidrule(lr){3-4}
    && FID$_{\downarrow}$ & KDR$_{\downarrow}$ \\
    \midrule
    \multirow{3}{30mm}{\textit{Joint Dataset} \\
    \textit{(Vox, AnimeCeleb)}}
    & FOMM & 100.95 & N/A \\
    & PIRenderer + $\mathcal{T}$ & 49.55 & N/A \\
    & Animo & 28.69 & 0.466 \\
    \midrule
    \multirow{1}{*}{\textit{AnimeCeleb}}
    & Ours & \textbf{23.15} & \textbf{0.236} \\
    \bottomrule
    \end{tabular}
    \caption{Quantitative results of animation face reenactment.
    FOMM~\cite{firstorder} and PIRenderer~\cite{ren2021pirenderer} are not available for KDR because distorted output makes keypoint detection impossible.}
    \vspace{-0.3cm}
    \label{Table:quantity}
\end{table}


\begin{table}[t!]
    \small
    \centering
    \begin{tabular}{ccc|c}
    \toprule
    \multicolumn{3}{c|}{\textbf{Loss}} & \multirow{2}{*}{\textbf{KDR$_{\downarrow}$}} \\
    \cmidrule(lr){1-3}
    Vertex & Landmark & E\&M Dist. &  \\
    \midrule
    \checkmark &  &  & 0.2431\\
     & \checkmark &  & 0.2394\\
    \checkmark & \checkmark &  & 0.2390\\
    \checkmark & \checkmark & \checkmark & \textbf{0.2360}\\
    \bottomrule
    \end{tabular}
    \caption{Ablation study on the loss component. E\&M Dist. indicates eye and mouth closure loss.}
    \vspace{-0.5cm}
    \label{Table:talkinghead}
\end{table}

\section{Conclusion}
In this paper, we propose a novel cross-domain expression translation network to map the human expressions to the anime expressions.
We achieve a significant improvement in the performance of cross-domain neural talking heads by implementing a shared 3D vertex space as a learning proxy.
Our model's superiority is validated through both quantitative and qualitative evaluations. 
As a direction for future work, we aim to train the network on a dataset of human expressions, allowing the network to mapping function as an explicit semantic controller.

\bibliographystyle{IEEEbib}
\bibliography{reference}

\begin{thebibliography}{10}

\bibitem{firstorder}
Aliaksandr Siarohin, St{\'e}phane Lathuili{\`e}re, Sergey Tulyakov, Elisa
  Ricci, and Nicu Sebe,
\newblock ``First order motion model for image animation,''
\newblock {\em Proc. the Advances in Neural Information Processing Systems
  (NeurIPS)}, vol. 32, pp. 7137--7147, 2019.

\bibitem{onefreeview}
Ting-Chun Wang, Arun Mallya, and Ming-Yu Liu,
\newblock ``One-shot free-view neural talking-head synthesis for video
  conferencing,''
\newblock in {\em Proc. of the IEEE conference on computer vision and pattern
  recognition (CVPR)}, 2021, pp. 10039--10049.

\bibitem{ren2021pirenderer}
Yurui Ren, Ge~Li, Yuanqi Chen, Thomas~H Li, and Shan Liu,
\newblock ``Pirenderer: Controllable portrait image generation via semantic
  neural rendering,''
\newblock in {\em Proc. of the IEEE international conference on computer vision
  (ICCV)}, 2021, pp. 13759--13768.

\bibitem{drobyshev2022megaportraits}
Nikita Drobyshev, Jenya Chelishev, Taras Khakhulin, Aleksei Ivakhnenko, Victor
  Lempitsky, and Egor Zakharov,
\newblock ``Megaportraits: One-shot megapixel neural head avatars,''
\newblock in {\em Proceedings of the 30th ACM International Conference on
  Multimedia}, 2022, pp. 2663--2671.

\bibitem{zhang2023metaportrait}
Bowen Zhang, Chenyang Qi, Pan Zhang, Bo~Zhang, HsiangTao Wu, Dong Chen, Qifeng
  Chen, Yong Wang, and Fang Wen,
\newblock ``Metaportrait: Identity-preserving talking head generation with fast
  personalized adaptation,''
\newblock in {\em Proceedings of the IEEE/CVF Conference on Computer Vision and
  Pattern Recognition}, 2023, pp. 22096--22105.

\bibitem{burkov2020neural}
Egor Burkov, Igor Pasechnik, Artur Grigorev, and Victor Lempitsky,
\newblock ``Neural head reenactment with latent pose descriptors,''
\newblock in {\em Proceedings of the IEEE/CVF conference on computer vision and
  pattern recognition}, 2020, pp. 13786--13795.

\bibitem{kim2022animeceleb}
Kangyeol Kim, Sunghyun Park, Jaeseong Lee, Sunghyo Chung, Junsoo Lee, and
  Jaegul Choo,
\newblock ``Animeceleb: Large-scale animation celebheads dataset for head
  reenactment,''
\newblock in {\em Proc. of the European Conference on Computer Vision (ECCV)}.
  Springer, 2022, pp. 414--430.

\bibitem{bansal2018recycle}
Aayush Bansal, Shugao Ma, Deva Ramanan, and Yaser Sheikh,
\newblock ``Recycle-gan: Unsupervised video retargeting,''
\newblock in {\em Proceedings of the European conference on computer vision
  (ECCV)}, 2018, pp. 119--135.

\bibitem{song2021everything}
Linsen Song, Wayne Wu, Chaoyou Fu, Chen Qian, Chen~Change Loy, and Ran He,
\newblock ``Everything's talkin': Pareidolia face reenactment,''
\newblock {\em arXiv preprint arXiv:2104.03061}, 2021.

\bibitem{xu2022motion}
Borun Xu, Biao Wang, Jinhong Deng, Jiale Tao, Tiezheng Ge, Yuning Jiang, Wen
  Li, and Lixin Duan,
\newblock ``Motion and appearance adaptation for cross-domain motion
  transfer,''
\newblock in {\em European Conference on Computer Vision}. Springer, 2022, pp.
  529--545.

\bibitem{nagrani2017voxceleb}
Arsha Nagrani, Joon~Son Chung, and Andrew Zisserman,
\newblock ``Voxceleb: a large-scale speaker identification dataset,''
\newblock {\em arXiv preprint arXiv:1706.08612}, 2017.

\bibitem{feng2021learning}
Yao Feng, Haiwen Feng, Michael~J Black, and Timo Bolkart,
\newblock ``Learning an animatable detailed 3d face model from in-the-wild
  images,''
\newblock {\em ACM Transactions on Graphics (ToG)}, vol. 40, no. 4, pp. 1--13,
  2021.

\bibitem{li2017learning}
Tianye Li, Timo Bolkart, Michael~J Black, Hao Li, and Javier Romero,
\newblock ``Learning a model of facial shape and expression from 4d scans.,''
\newblock {\em ACM Trans. Graph.}, vol. 36, no. 6, pp. 194--1, 2017.

\bibitem{gerig2018morphable}
Thomas Gerig, Andreas Morel-Forster, Clemens Blumer, Bernhard Egger, Marcel
  Luthi, Sandro Sch{\"o}nborn, and Thomas Vetter,
\newblock ``Morphable face models-an open framework,''
\newblock in {\em 2018 13th IEEE International Conference on Automatic Face \&
  Gesture Recognition (FG 2018)}. IEEE, 2018, pp. 75--82.

\bibitem{huang2017arbitrary}
Xun Huang and Serge Belongie,
\newblock ``Arbitrary style transfer in real-time with adaptive instance
  normalization,''
\newblock in {\em Proceedings of the IEEE international conference on computer
  vision}, 2017, pp. 1501--1510.

\bibitem{anime-face-detector}
hysts,
\newblock ``Anime face detector,''
  \url{https://github.com/hysts/anime-face-detector}, 2021.

\end{thebibliography}

\end{document}